%% file: example_paper.tex
\documentclass[nohyperref]{article}

\usepackage{microtype}
\usepackage{graphicx}
\usepackage{subfigure}
\usepackage{booktabs} % for professional tables
\usepackage{listings}
\usepackage{tabularx, makecell, multirow} 
\usepackage{pythonhighlight}

\newcommand\figref[1]{Figure~\ref{#1}}
\newcommand\tabref[1]{Table~\ref{#1}}
\newcommand\secref[1]{Section~\ref{#1}}

\usepackage{hyperref}

\usepackage[accepted]{icml2022}

\usepackage{amsmath}
\usepackage{amssymb}
\usepackage{mathtools}
\usepackage{amsthm}

\usepackage[capitalize,noabbrev]{cleveref}

\theoremstyle{plain}

\theoremstyle{definition}

\theoremstyle{remark}

\usepackage[textsize=tiny]{todonotes}

\icmltitlerunning{Efficiera Residual Networks}

\begin{document}

\twocolumn[
\icmltitle{Efficiera Residual Networks: Hardware-Friendly Fully Binary Weight with 2-bit Activation Model Achieves Practical ImageNet Accuracy}

\icmlsetsymbol{equal}{*}

\begin{icmlauthorlist}
\icmlauthor{Shuntaro Takahashi}{takahashi}
\icmlauthor{Takuya Wakisaka}{wakisaka}
\icmlauthor{Hiroyuki Tokunaga}{tokunaga}
\end{icmlauthorlist}

\icmlaffiliation{takahashi}{This work was performed while the 3 authors were affiliated with LeapMind Inc. Currently, Shuntaro Takahashi is affiliated with M3, Inc.}
\icmlaffiliation{wakisaka}{Takuya Wakisaka is currently a freelance engineer/researcher. \textless{}takuya.wakisaka@moldweorp.com\textgreater{}.}
\icmlaffiliation{tokunaga}{Hiroyuki Tokunaga is currently a freelance engineer/researcher. \textless{}tokunaga.hiroyuki@gmail.com\textgreater{}.}

\icmlcorrespondingauthor{Shuntaro Takahashi}{shushakura@gmail.com}

\icmlkeywords{classification, ultra-low-bit quantization, edge-device inference}

\vskip 0.3in
]
\printAffiliationsAndNotice{}  % leave blank if no need to mention equal contribution

\begin{abstract}
The edge-device environment imposes severe resource limitations, encompassing computation costs, hardware resource usage, and energy consumption for deploying deep neural network models. Ultra-low-bit quantization and hardware accelerators have been explored as promising approaches to address these challenges. Ultra-low-bit quantization significantly reduces the model size and the computational cost. Despite progress so far, many competitive ultra-low-bit models still partially rely on float or non-ultra-low-bit quantized computation such as the input and output layer. We introduce Efficiera Residual Networks (ERNs), a model optimized for low-resource edge devices. ERNs achieve full ultra-low-bit quantization, with all weights, including the initial and output layers, being binary, and activations set at 2 bits. We introduce the \textit{shared constant scaling factor} technique to enable integer-valued computation in residual connections, allowing our model to operate without float values until the final convolution layer. Demonstrating competitiveness, ERNs achieve an ImageNet top-1 accuracy of 72.5pt with a ResNet50-compatible architecture and 63.6pt with a model size less than 1MB. Moreover, ERNs exhibit impressive inference times, reaching 300FPS with the smallest model and 60FPS with the largest model on a cost-efficient FPGA device. The model implementation is available \href{https://github.com/LeapMind/ERN}{here}.
\end{abstract}

\section{Introduction}
The computational cost of deep neural networks has been a bottleneck in deploying the models to the low-resource environment. The hardware accelerator and the model quantization have been intensively investigated to realize the deployment. The low-bit quantization of deep learning models reduces the computational cost drastically. Ultra-low-bit quantization reduces the weight and activation to binary, fractional-bits, ternary or 2-bits \citep{hubara2016, rastegari2016, zhang2018, liu2018, liu2020, bethge2021, zhang2022, tu2022, guo2022, zhang2021, wan2018, zhou2016, cai2017}. While early works suffered from the huge gap with float counterparts, recent ultra-low-bit quantized models successfully mitigated the accuracy loss. They achieved equivalent classification accuracy of ResNet-18 / MobileNetV2 on a large-scale dataset such as ImageNet with a small model size compared to float models \citep{liu2020,zhang2022, guo2022}.

\begin{figure}[t]
    \centering
    \includegraphics[width=90mm]{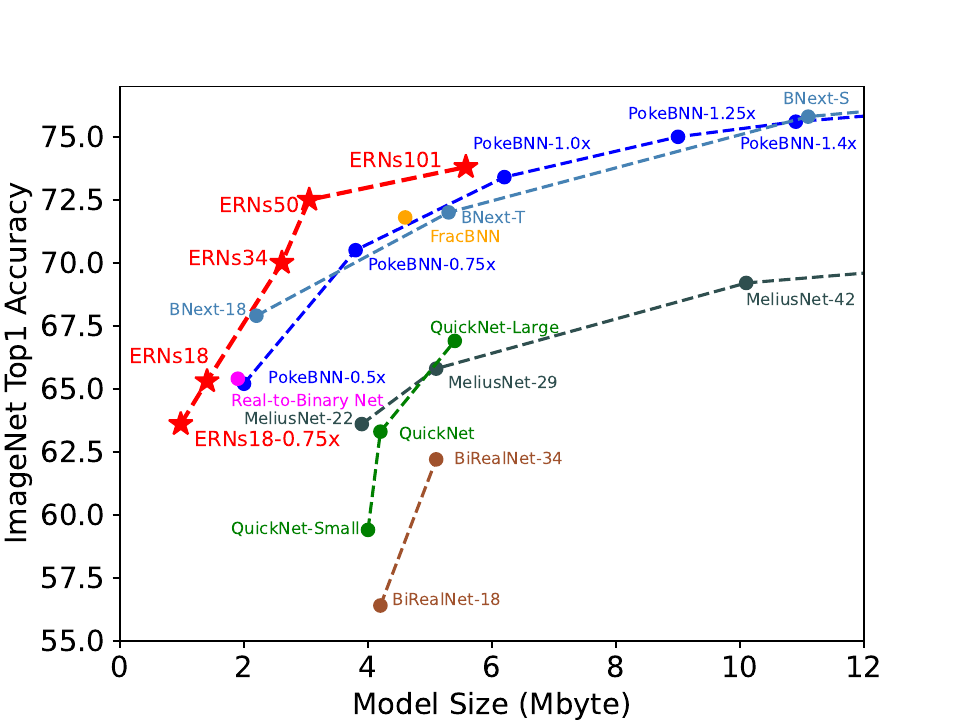}
    \caption{Model Size and ImageNet Top-1 Accuracy of the ultra-low-bit quantized models. See also \secref{sec:result}.}
    \label{fig:size_accuracy}
\end{figure}

While significant progress has been made in years, existing BNNs still involve float or int8-quantized convolutional neural networks (CNNs) to prevent a radical accuracy drop. \tabref{tab:taxonomy} summarizes the use of weight-activation bit pairs, variations of CNNs, and non-CNN operations used in the ultra-low-bit quantized models.
\input{./table/taxonomy_table}
Typically, the input and output layers of the competitive ultra-low-bit quantized models are int8-quantized or float16 to mitigate the accuracy drop. Some recent work introduces advanced network designs using multiple bit-level quantization such as the down-sampling by the average pooling with float 1x1 convolution \citep{liu2018,liu2020}, 8-bit depth-wise separable convolution (DWConv) and 4-bit squeeze-and-excitation (SE) structures \cite{zhang2022} to lessen the performance degradation due to the ultra-low-bit quantization with a carefully designed small fraction of non-low-bit computation. They successfully improve the accuracy with a minimal increase in the cost of arithmetic computation.

However, from the hardware accelerator design's point of view, non-low-bit or even float-valued computation and variations of CNNs such as DWConv and SE should be avoided for the following reasons.

Firstly, the support of various operations and different-bit pair computation on the hardware accelerator requires the implementation of additional arithmetic circuits. It leads to an increase in the circuit area for ASIC or the use of FPGA resources regardless of their computational cost. For the existing ultra-low-bit quantized models, because the input and output layers are 8-bit or float, they need an additional 8-bit or float CNNs execution circuit or the operations are outsourced to the CPU \cite{zhang2021}.

Secondly, if the model involves operations that can not be executed on the hardware accelerator, it needs to be executed on the CPU and data transfer between the CPU and hardware accelerator occurs. It consumes the DRAM bandwidth and may become the bottleneck of the inference time.

Thirdly, the variations of CNN architectures such as DWConv and SE is likely to decrease the utilization of the hardware accelerator compared to the theoretical arithmetic operation cost. Even within the hardware accelerator, the data transfer is more likely to be the bottleneck than the arithmetic operation \cite{radosavovic2020}. The support of such variations may just increase the circuit area and does not contribute to the reduction of inference time.

Based on these insights, we consider that a hardware-friendly model should satisfy the following conditions.
\begin{enumerate}
\item The weight and activation bits should be low and the pair of them should be only one. This policy applies to all layers including the input and output layers. In this paper, we select binary weight and 2-bit activation.
\item The standard CNNs should only be used and the use of advanced CNNs such as DWConv and SE and non-CNN operations such as max pooling and average pooling should be avoided in-between a sequence of CNN layers.
\item The float-valued computations such as the addition of feature maps of residual connection should be removed.
\end{enumerate}

To develop the model satisfying these conditions, we design novel network modules as follows. 
\begin{itemize}
    \item \textit{pixel embedding} module: The original 8-bit pixel values need to be transformed into 2-bit activation maps to make the first conv layer w1a2. We introduce \textit{generalized thermometer encoding} based on the idea of thermal encoding \cite{buckman2018, zhang2021} to effectively represent the pixel values in lower-bits .
    \item stem module: Unlike ResNet or MobileNet, the max pooling layer should be replaced with the conv layer to down-sample the \textit{pixel embedded} input. We design the simple stem module for ERNs. 
    \item integer-valued residual connection: Even with ultra-low-bit quantized models, the residual connection is computed as the addition of float values due to the presence of scaling factor. We introduce a novel technique \textit{shared constant scaling factor} to allow to compute the residual connection as addition of integer values.
    \item classifier module:  The de facto standard design of classifier module applies the average pooling first to reduce the spatial dimensions and linear classifier afterward. We reverse the order of these operations for ERNs to make the network architecture so simple that it consists of the repeats of conv-bn-act with the integer-valued residual connection from the stem module to the final conv layer. 
\end{itemize}

We demonstrate our ERNs achieve competitive accuracy on ImageNet. \figref{fig:size_accuracy} shows the scatter plot of the model size and top1 accuracy on ImageNet of ultra-low-bit quantized models. As all parameters of ERNs are binary-valued, the model size is significantly smaller than the other comparable quantized model. We develop multiple models including the tiniest model whose weight is less than 1 megabyte and large ResNet-101 compatible model whose weight is only 5.6 megabyte. The accuracy reaches over 70.0 pt accuracy with ResNet-50 compatible model with only 3.1 mega bytes. Notably, our training scheme remains simple as it does not involve two-phase training or knowledge distillation. 

Thanks to the hardware-friendly network design, ERNs can be executed quickly with the cost efficient FPGA device. We demonstrate the inference on the cost-efficient FPGA device Kria KV260. Our tiniest model runs at 298.5 FPS and the heaviest model runs at 63.4 FPS with our IP called \textit{Efficiera}. By fully removing the non-low-bit computation, \textit{Efficiera} does not use any DSP blocks.
\section{Related Works}
\label{sec:related}
\subsection{Low-bit quantization of neural networks}
From the pioneering works \cite{hubara2016, cai2017}, the ultra-low-bit quantization has been investigated extensively and the model accuracy has improved by far to close the gap with the float counterpart. 
There have been several approaches such as the quantization method \cite{rastegari2016,zhou2016,cai2017,zhang2018,esser2020}, the design of network architecture \cite{liu2018, bethge2021, zhang2022,guo2022}, and the new training scheme \citep{liu2020, tu2022}.

As highlighted in \tabref{tab:taxonomy}, a common practice in ultra-low-bit quantized models involves maintaining float or int8-quantized input and output layers, recognizing the substantial accuracy drop associated with ultra-low-bit quantization in these layers.

For the input layer, \citet{durichen2018} proposed to use fixed point integer to binarize the input. \citet{zhang2021} used the binary weight CNNs and thermometer encoding, which transforms the 8-bit input into a vector of binary values and achieved 71.8pt on ImageNet. \citet{tokunaga2024} proposed Pixel Embedding, which utilizes the learnable embedding table to quantize the input. For the output layer, \citet{hubara2016} and \citet{nakahara2017} used the binary-weighted classifier, yet their model's accuracy was quite far from ResNet18 accuracy. 

Recent works introduce the fine-grained designs of network blocks to enhance the ultra-low-bit quantized models. \citet{liu2018} proposed Bi-Real Net, which uses the float-valued 1x1 conv in the down-sampling layers. \citet{zhang2022} designed PokeInit and PokeConv modules, which carefully introduced 4-bit and 8-bit quantized DWConv and SE architectures. \citet{guo2022} proposed Info-RCP module to enhance the capacity by SE architecture with w4a8 quantization. These efforts are effective in improving the accuracy with a small arithmetic computation cost of the models, however, it should also be noted that the complex network design and non-low-bit computation lead to the necessity of a model-specific accelerator design and may lead to the increase of the circuit area.

\subsection{Network Design Optimization for Computer Architectures}
Due to the diverse needs for deep learning models, there have been multiple platforms deployed such as the CPU, mobile GPU, FPGA, and ASIC. As the characteristics of them are different, the network design for platforms has been explored in recent years.

MobileNet \cite{sandler2018}, for example, was designed primarily for mobile processors. Because the mobile processors have insufficient processing speed, the number of operations is likely to be the bottleneck of the inference time. MobileNet drastically reduced the number of operations by using DWConv. SE block \cite{hu2018} is another method to improve the accuracy with a minimal increase of the number of operations.

\citet{radosavovic2020} pointed out that the operation reduction with DWConv only slightly contributes to the inference time on GPU and other hardware accelerators. They observed that the inference time strongly correlates with activations, the size of the output tensors of all CNN layers. For the hardware accelerators, the convolution operation can be executed quickly, but the data transfer of output tensors is more likely to be a bottleneck to the inference time. They designed the search space of the network architectures based on activations and obtained high-accuracy models with shorter inference time.

\citet{yang2018} analyzed the computational cost of the residual connection for FPGA device with 4-bit weight and activation model. They argued that concatenative connection requires smaller data movement than additive connection. It seems, however, difficult for ultra-low-bit quantized models to achieve competitive accuracy without additive connection as existing competitive models rely on it.

\citet{guo2021} investigated the energy consumption of the BNNs hardware accelerator by writing RTL and pointed out that the float-valued computations degrade the energy efficiency. They proposed BoolNet, in which they radically reduced the representation of feature maps to 1/4 bits integer to reduce the energy consumption caused by data transfer.

\citet{zhang2021} had the most similar motivation with us and implemented the ultra-low-bit quantized model named FracBNN. It can be effectively executed on the FPGA device and achieved the MobileNet-V2 equivalent ImageNet accuracy. They carefully investigated the impact of bit-width changes on FPGA logic usage. Under their setting, the replacement of the input layer from \textit{w1a1} to \textit{w8a8} increases the DSP logic unit from 3 to 287, BRAM from 2.5 to 34, and LUTs from 3603 to 22509. This significant difference in logic usage raises the necessity to remove the non-low-bit CNNs and use only solo bit-pair.

\section{Efficiera Residual Networks}
\subsection{Ultra-low-bit quantization of CNNs}
Convolution operation produces the feature map output $\mathbf{O}$ from the computation defined between the weight parameters $\mathbf{W}$ and the inputs $\mathbf{X}$

\begin{equation}
    \mathbf{O} = \mathbf{W} \ast \mathbf{X}.
\end{equation}

For ultra-low-bit quantized CNNs, the weights and activated inputs are quantized and a scaling factor $\alpha$ \cite{cai2017} is applied  
\begin{equation}
    \mathbf{\tilde{O}} = \alpha(\mathbf{\tilde{W}} \ast \mathbf{\tilde{X}}),
\end{equation}
where $\mathbf{\tilde{W}}$ and $\mathbf{\tilde{X}}$ are the quantized weights and inputs. The feature map $\mathbf{\tilde{O}}$ is usually float-valued due to the presence of the scaling factor $\alpha$.  It is consumed by the batch normalization and activation function to be quantized and becomes the input $\mathbf{\tilde{X}}$ of the following CNNs layer. 

We use the notion $l$ to denote the number of bits. For binary quantization of weights, the sign function is applied to convert the values into 1 and -1
\begin{equation}
    \mathtt{sign}(x)=\begin{cases}
+1,& \mathrm{if} \ x \ge 0,\\
-1,& \mathrm{otherwise}.
\end{cases}
\end{equation}
The scaling factor $\alpha$ can be a constant, computed from the norm of float weight $\mathbf{W}$ \cite{rastegari2016}, or  a learnable parameter  \cite{esser2020}.

For the quantized activation function $q(x)$, ReLU-like non-linearity \cite{cai2017} is used in most ultra-low-bit quantized models. With the most simple case, the input scalar $x$ is clamped between $0$ and $2^{l}-1$ and the floor function is applied
\begin{equation}
    q(x) = \text{floor}(\text{clamp}(x, 0, 2^{l}-1)).
\end{equation}
In practice, the activation function also involves a scaling factor for improving accuracy \cite{esser2020}.

\subsection{Network Architecture Design with Minimal Operation Set}
We design our ERNs to keep the network structure and operations simple. Specifically, our network consists of the following operations.
\begin{itemize}
    \setlength\itemsep{-0.5em}
    \item pixel embedding module with \textit{generalized thermometer encoding}
    \item \textit{w1a2} CNNs
    \item batch normalization and 2-bit quantized activation function
    \item addition of integer-valued feature maps
    \item average pooling at the very end of the network architecture
\end{itemize}
We do not use advanced CNNs architectures such as DWConv and SE because they cause an increase in the circuit area and their computational efficiency tends to remain low. We locate the average pooling layer at the very end of the network structure so that the computation except the final pooling operation is executed only by the accelerator. We also limit the output channel sizes of the CNNs layer to a multiple of $64$. This limitation is beneficial for the accelerators to enhance the parallelism of the computation \citep{jouppi2017, kung2019, zhang2021}. It should be mentioned that this limitation and the removal of the max pooling layer from the stem module inevitably increase the number of operations compared to the original ResNet and derived ultra-low-bit quantized models. In a later experiment, we demonstrate that the inference time on the cost-effective FPGA device is competitive and even superior to the existing works.

\begin{figure}[t]
    \centering
    \includegraphics[width=80mm]{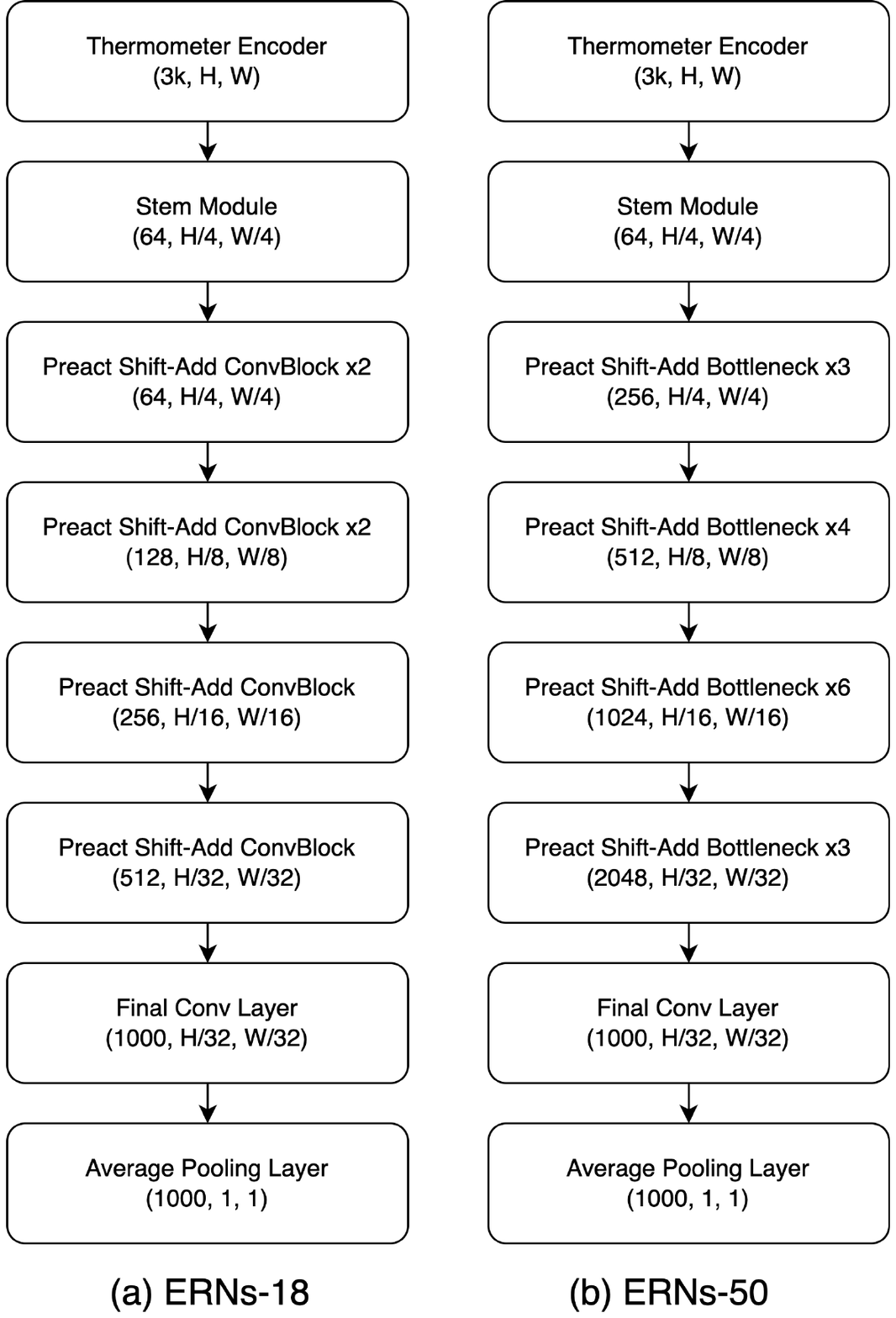}
    \caption{Network architecture of ERNs. The values in the bracket are the output shape in (channel, height, width).}
    \label{fig:architecture}
\end{figure}

The overall network structure of our model is shown in \figref{fig:architecture}. The macroscopic design is inspired by ResNet \cite{he2016}. The stem module reduces the space dimensions by 4, and repeats of CNNs with residual connections are applied to amplify the channel size and reduce the space dimensions by 32. In the following, we explain the details of our network design including the input pixel embedding module, the network structure of the stem module, the design of the residual connection, and the output layer.

\subsection{Pixel Embedding Module: Generalized Thermometer Encoding}
To low-bit quantize the first CNNs layer computation, the input 8-bit-valued pixels should be embedded into a low-bit representation instead of the standard float-valued normalization. We refer to this operation as \textit{pixel embedding}. Thermometer encoding was first proposed in the context of adversarial attack \cite{buckman2018} and applied in low-bit quantization in \citet{zhang2021}. In their context, thermometer encoding appeared as a method to represent an integer as a binary-valued vector. As our model is \textit{w1a2}, we generalize it to $l$-bit and use $l=2$.

Intuitively, thermometer encoding converts an 8-bit integer into a channel-wise monotonically increasing fixed-length lower $l$-bit vector. For example, if the vector length is $2$ and has 2-bit values, an 8-bit integer is evenly mapped to one of $00$, $01$, $11$, $12$, $22$, $23$, and $33$ according to the magnitude of the integer in monotonic order. Unlike the base-4 positional numeral system, thermometer encoding preserves the order of magnitude for each channel.

We formulate the \textit{generalized thermometer encoding} as follows. Let $x$ be an 8-bit integer and $\mathbf{z}$ the corresponding $l$-bit thermometer-encoded vector with a length of $k$. The vector $\mathbf{y}$ of length $k$ is defined as the linear function 
\begin{equation}
    \mathbf y = \mathbf{w}x + \mathbf b
\end{equation}
The vectors $\mathbf{w}$ and $\mathbf{b}$ also have a length of $k$ and the $i$-th element of the vector $\mathbf{w}$ and $\mathbf{b}$  are defined as follows.  
\begin{eqnarray}
    \mathbf{w}_{i} & = & \frac{1}{sk}, \\
    \mathbf{b}_{i} & = & 1-\frac{i+1}{k}, \\
    s & = & \max(1, \lfloor\frac{255}{(2^{l}-1)k} \rfloor).
\end{eqnarray}

The parameter $s$ is the bin-width of a unique $l$-bit vector. The floor function and clamp functions are finally applied to obtain the $l$-bit integer-valued vector $\mathbf{z}$.
\begin{equation}
\mathbf{z}_{i} = \text{clamp}(\lfloor \mathbf{y}_{i} \rfloor, 0, 2^{l}-1).
\end{equation}

\subsection{Stem Module}
The pixel-embedded image representation is down-sampled by the stem module, which reduces the dimensions of the space by $4$ and outputs $64$ channels following ResNet. The original ResNet employs 7x7conv for the first layer with a stride of 2 and applies the max pooling operation with a kernel size of 3x3.  As discussed in, the first layer of ResNet requires 118 million MACs with 224x224 input due to the large kernel size and the large spatial resolutions. In our case, because the input channel is 30 instead of 3 due to the pixel embedding, the number of operations becomes $10$ times larger, 1.18 billion MACs.

Although it is inevitable to increase the arithmetic computational cost by replacing the max pooling with the down-sampling CNN layer, we lessen its increase by removing the large kernel like 7x7. Our stem module consists of 4 layers of 3x3conv with $64$ output channels and the first and third layers are stride 2 to reduce the spatial dimensions by 4.

% \begin{figure}[ht]
% \begin{python}
% def forward(self, x):
%   x = conv3x3(x, stride=2, bn_act=False)
%   x = conv3x3(x, stride=1, bn_act=True)
%   x = conv3x3(x, stride=2, bn_act=True)
%   x = conv3x3(x, stride=1, bn_act=True)
%   return x
% \end{python}
% \caption{Stem module architecture. The output channel is $64$ for all layers.}
% \label{code:stem}
% \end{figure}

\subsection{Integer-Valued Residual Connection}
The residual connection \cite{he2016} is a key to train deep neural networks. Successful ultra-low-bit quantized models also adopt residual network architectures. The calculated feature maps of ultra-low-bit quantized model $\mathbf{\tilde{O}}$ are float-valued due to the presence of scaling factor $\alpha$. The residual connection consequently requires the addition of the float-valued feature maps, which demand a non-negligible computational cost compared to low-bit addition. According to \cite{jouppi2021}, the addition of float32 consumes roughly 13 times more energy than that of int32.

To replace the float-valued addition of feature maps with an integer-valued counterpart, we propose \textit{shared constant scaling factor}. The idea is to fix the scaling factors before the residual connection the same constant so that the computation of the scaling factor can be delayed after the addition of the integer valued feature maps. With further optimization of the computation, the multiplication of the scaling factor $\alpha$ can be fused into the quantized activation functions by changing their threshold \citep{zhang2018, leapmind2021}. Therefore, the float-valued computation never appears during inference on the hardware accelerator as the integer-valued feature maps are directly transformed to 2-bit values including the residual connection.

\figref{code:block} and \figref{code:bottleneck} present the python code for ConvBlock and Bottleneck block respectively. The parameters \pyth{post_sc} and \pyth{const_sc} stand for the post scaling and constant scaling. The inputs of the residual connection are guaranteed to be integer-valued by disabling the scaling factor of the activation function and setting the shared constant scaling factor to the weight quantizer.
\begin{figure}[ht]
\begin{python}
def forward(self, x, downsample=True):
  identity = x
  out = bn_0(x)
  out = act_0(x, post_sc = not downsample)
  if downsample:
    identity = down_conv(out,const_sc=True)
  out = conv_1(out)
  out = bn_1(out)
  out = act_1(out, post_sc=False)
  out = conv_2(out, const_sc=True)
  out += identity
  return output
\end{python}
\caption{ConvBlock architecture for ERNs}
\label{code:block}
% \end{figure}
% \begin{figure}[ht]
\begin{python}
def forward(self, x, downsample=True):
  identity = x
  out = self.bn_0(x)
  out = act_0(x, post_sc = not downsample)
  if self.downsample:
    identity = down_conv(out)
  out = self.conv_1(out)
  out = self.bn_1(out)
  out = self.act_1(out)
  out = self.conv_2(out)
  out = self.bn_2(out)
  out = self.act_2(out,post_sc=False)
  out = self.conv_3(out,const_sc=True)
  out += identity
  return out
\end{python}
\caption{Bottleneck architecture for ERNs}
\label{code:bottleneck}
\end{figure}

\subsection{Output Layer}
The final conv layer of our model is 1x1 conv with a binary weight and a scaling factor. It is a challenge to mitigate the accuracy drop compared to float or int8-quantized classifiers. The average pooling operation is applied after the final conv layer to reduce the spatial dimensions. This design is intended to avoid non-CNN layers in-between CNN layers to maximize the utilization of the hardware accelerator and reduce the data transfer with CPU.

\section{Experiment}
\subsection{Experiment Settings}
In the experiment, we set up ERNs18-x0.75, ERNs-18, ERNs-34, ERNs-50, and ERNs-101. The later 4 models have the comparable ResNet models in terms of the number of layers, channel sizes, and the residual connection blocks. ERNs18-x0.75 reduces the channel size of last blocks from 512 to 384 to make the model size under 1 megabyte. To the best of the authors' knowledge, this is the first ultra-low-bit quantized model that has the model weight less than 1 megabyte and achieves practical accuracy.

For quantization, except for the \textit{shared constant scaling factor} for residual blocks, the weights are binarized with a scaling factor proposed in  \cite{rastegari2016} and the activations are ReLU function with the floor function \cite{cai2017}. We use the element-wise gradient scaling with the hyper-parameter $\epsilon=0.01$ \cite{lee2021}. 

We use Adam optimizer \cite{kingma2014} with a base learning rate of $4\cdot10^{-4}$ for batch size $256$. We linearly scaled the learning rate according to the batch size. The hyperparameter $\epsilon$ of the Adam optimizer is set to $1\cdot10^{-4}$ . We set the training epochs 400 and apply a cosine learning rate scheduler. A weight decay of $1\cdot10^{-5}$  is applied. RandAugment \cite{cubuk2020} is employed for the training data augmentation with an input image size of 256x256. The batch normalization momentum is set to $0.9$. 

For evaluation, we use 2 techniques deviated from the standard single inference. Firstly, we use an increased resolution from training to improve the accuracy. This phenomenon is known as ``train-test resolution discrepancy'' \cite{touvron2019}. Secondly, we use ten-crop test-time augmentation to ensemble the inference result from multiple crops, which has been employed in AlexNet \cite{krizhevsky2012}, ResNet \cite{he2016}, and many other works. We averaged the logit outputs first and computed the class probability by softmax.

\subsection{Result}
\label{sec:result}
\input{./table/ern_result_table}
As summarized in \tabref{tab:accuracy}, our ERNs achieve higher accuracy with a smaller model size compared to the previous ultra-low-bit quantized models. For fairness, it should be mentioned that the number of operations of ERNs is larger than models derived from ResNet mainly due to the replacement of the max pooling layer with CNN layers in the stem module. 

\input{./table/gops}

\tabref{tab:gops} compares the number of operations of ERNs and ResNet models with 256x256 input.

We evaluate the effect of the inference scheme differences from the training.
\tabref{tab:resolution} shows the effect of the input resolution at inference time. We observed that the accuracy reaches the maximum with 288x288 input while the model is trained with 256x256 input resolution. Increasing the input resolution roughly multiplies the number of arithmetic computations by 1.3 times. The ten-crop augmentation contributed about 2 pt increase.

\input{./table/resolution}

We introduced \textit{generalized-thermometer encoder}, which is different from the standard float-valued input with normalization. \tabref{tab:thermoencoder} shows the effect of \textit{generalized-thermometer encoder}'s $k$ to the accuracy. The results are with ERNs-18 and the training epoch is reduced to 100. While increase of $k$ from 10 to 21 improves the accuracy by 0.7 points, the increase from 21 to 32 does not improve the accuracy. Considering the trade-off between the gain of the accuracy and the increase of the arithmetic computation cost. The value $k=10$ is sufficiently large enough to obtain the competitive accuracy.

Overall, ERNs achieved the competitive and practical accuracy on ImageNet dataset even it has unfavorable conditions such as the ultra-low-bit quantized input/output layers, information-lossy pixel embedding with \textit{generalized-thermometer encoding}, and the restriction in the residual connection with \textit{shared constant scaling factor}. They are introduced to make the inference faster on the edge devices. In \secref{sec:deployment}, we show the outperformed inference speed on the FPGA device with smaller logic use.  

\begin{table}[]
\caption{The effect of thermometer encoder's $k$ to the top-1 accuracy of ImageNet. The accuracy is lower than values reported in \tabref{tab:accuracy} because the training epoch is reduced from 400 to 100.}
\label{tab:thermoencoder}
\begin{center}
\begin{tabular}{l|c c c}
\hline
Thermometer Encoder's $k$ & 10 & 21 & 32 \\ \hline
ERNs-18 & 57.4 & 58.1 & 58.0 \\ \hline
\end{tabular}
\end{center}
\end{table}

\section{Deployment to FPGA Device}
\label{sec:deployment}
We present the inference performance on the FPGA device. We use a Kria KV260 with Zynq UltraScale+ MPSoC device from Xilinx Inc.
Our FPGA design is \textit{Efficiera} IP developed by LeapMind Inc.. \textit{Efficiera} IP supports the operation sets for ERNs except for the average pooling at the very end of the network architecture.

\tabref{tab:fpga} compares the FPGA resource usage of our \textit{Efficiera} with existing works.
DPU B4096 is the IP released as a part of Vitis-AI from Xilinx Inc. that supports the inference of 8-bit quantized models.
Notably, our configuration does not use any DSP block. As the number of DSP blocks is quite limited for cost-efficient FPGA devices, this is a huge advantage over other implementations. It is also favorable for end-users of FPGA devices because they are also interested in implementing classical image processing modules on FPGA requiring DSP blocks.

\tabref{tab:kv260} shows the inference time of ERNs on Kria KV260 with \textit{Efficiera} IP. We measure the inference time with the first 1,000 images of the validation subset of ImageNet dataset. The input resolution is fixed to 256x256 and the batch size is 1.

Because our network can fully utilize the hardware accelerator from the very beginning to the end except for the final average pooling, the inference time is quite fast and outperforms existing works. For the tiniest ERNs18-0.75x, the single inference runs only within 3.35ms. Even for the largest ERNs101 compatible with ResNet101, it can be executed within 15.78 ms.

Compared to the existing comparable models, the number of operations of ERNs is larger and this is superficially a drawback. However, the inference time of ERNs with \textit{Efficiera} is faster than existing models with a smaller number of operations. On hardware accelerators, the cost of data transfer overwhelms that of arithmetic computation because arithmetic computation can be effectively parallelized while data transfer does not. Therefore, the number of activations is a more meaningful metric to estimate the inference time from network architecture \cite{radosavovic2020}. ERNs demonstrated that the hardware-friendly minimal operations set is sufficient to achieve competitive accuracy and fast inference time.

\input{./table/fpga}
\input{./table/kv260}
\section{Discussion}
\textbf{Comparison with Recent Binary Neural Networks}

In recent years, the accuracy of binary neural networks (BNNs) have been improved, which has binary weight and activation. The pair of binary weight and activation is the most promising because it can be computed by a single logic gate. Yet, with respect to the progress of BNNs, it is infeasible to develop a high-accuracy model with fully binarized weights and activations.

As mentioned in \secref{sec:related} and summarized in \tabref{tab:taxonomy}, the recent architectural improvements such as Real-to-Binary Net, ReActNet, and PokeBNN involve float-valued or non-low-bit quantized operations.
If the model is executed on CPU with the inference framework such as BitFlow \cite{hu2018bitflow}, Larq \cite{geiger2020}, and dabnn \cite{zhang2019}, these non-low-bit quantized and float-valued computations are not huge burden because CPU supports the execution of them.

However, if we consider to run the inference under the severe resource limitations such as the number of logic units and power supply, we need to implement the hardware accelerators with either FPGA or ASIC. If these advanced modules need to be supported by the hardware accelerator such as FPGA, the additional logic units should be implemented even the ratio of the computational cost is insignificant. Therefore, it is not always fair to compare the inference cost by the arithmetic computation cost \cite{zhang2022} if the model is designed for the hardware accelerators.

In this paper, we demonstrated that the fast inference can be realized with the cost-efficient FPGA device even the arithmetic computation cost increased compared to BNNs.

\textbf{Radical model size reduction by full binarization}

Reducing the size of the model is critical for hardware accelerators. On the hardware accelerator, weights should be located on SRAM, which is often mounted only a limited amount due to the financial cost.
The final conv layer of ResNet18 and ResNet50 has 512,000 and 2048,000 parameters for ImageNet classification. If the parameters are float16, they require 1 and 4 megabytes. By binarizing these parameters, they are reduced to 0.061 and 0.244 megabytes. It makes us possible to develop the hardware-friendly model under 1.0 megabyte.

\section{Conclusion}
In conclusion, we introduced Efficiera Residual Networks (ERNs), a novel approach characterized by fully binary-weighted architecture with 2-bit activation. Our contributions include the innovative \textit{Generalized-thermometer encoding} method for pixel-embedding to enable ultra-low-bit quantization of the input convolutional layer and the \textit{Shared constant scaling factor} to ensure integer-valued computation in the residual connection.

In comparison to existing ultra-low-bit quantized models, ERNs exhibit a reduced model weight size and demonstrate effective optimization for hardware accelerators. Notably, our model successfully mitigates the accuracy drop associated with all-layer quantization, achieving accuracy levels comparable to float-based ResNet-18 with ERNs-50, maintaining compatibility with ResNet-50 architecture.

For future work, we envision enhancing training schemes, particularly exploring knowledge distillation and multi-step training strategies. These endeavors aim to further refine the model's performance and contribute to the ongoing advancement of ultra-low-bit quantized neural networks.

\newpage
\bibliography{example_paper}
\bibliographystyle{icml2022}
\newpage
\appendix
\onecolumn

\end{document}

%% file: table/taxonomy_table.tex
\begin{table*}[t]
\setlength{\tabcolsep}{4pt}
\caption{The taxonomy of ultra-low-bit quantized network architectures. The existing competitive models involve non-ultra-low-bit computations in input layers, classifiers, and even the intermediate CNNs layers. Our model removes non-low-bit CNNs so that the hardware only needs to support a certain bit-pair CNNs add feature maps for residual connection. $\ast$: FracBNN uses w1a1.4 fractional activation, computed as 2-time w1a1 CNNs. {\S}: t in  w1at stands for the ternary activation.}
\label{tab:taxonomy}
\begin{center}
\begin{footnotesize}
\begin{tabular}{l|c|c|c|c|c|c}
\Xhline{2\arrayrulewidth}
model  &  \begin{tabular}{c} input layer \\ bit-pair\end{tabular}&\begin{tabular}{c} intermediate layer \\ bit-pair\end{tabular}&\begin{tabular}{c} classifier \\ bit-pair\end{tabular}& CNNs type& \begin{tabular}{c} residual \\ connection\end{tabular} & non-CNNs layers\\ \Xhline{2\arrayrulewidth}
BNNs &  w1af&w1a1&w1a1& normal  & float  & \begin{tabular}{c} maxpool (stem) \\ avgpool (intermediate)\end{tabular}\\ \hline
\noindent\begin{tabular}{@{}l}XNOR-Net,Bi-RealNet\\MeliusNet, ReActNet\end{tabular} & wfaf&w1a1, wfaf &wfaf& normal  & float  & \begin{tabular}{c} maxpool (stem) \\ avgpool (intermediate)\end{tabular}\\ \hline
\noindent\begin{tabular}{@{}l}DoReFa-Net\\LQ-Net, HWGQ\end{tabular} &  wfaf&w1a2&wfaf& normal  & float  & \begin{tabular}{c} maxpool (stem) \\ avgpool (intermediate)\end{tabular}\\ \hline
% LQ-Net &  wfaf&w1a2&wfaf& normal  & float  & \begin{tabular}{c} maxpool (stem) \\ avgpool (intermediate)\end{tabular}\\ \hline
% HWGQ       &  wfaf&w1a2 &wfaf& normal & float  & \begin{tabular}{c} maxpool (stem) \\ avgpool (intermediate)\end{tabular}\\ \hline
% Bi-RealNet &  wfaf&w1a1, wfaf  &wfaf& normal & float  & \begin{tabular}{c} maxpool (stem) \\ avgpool (intermediate)\end{tabular}\\ \hline 
Real-to-Binary Net& wfaf& w1a1, wfaf & wfaf & \begin{tabular}{c} normal \\ SE   \end{tabular}& float & \begin{tabular}{c} maxpool (stem) \\ avgpool (intermediate)\end{tabular}\\\hline
% MeliusNet  &  wfaf&w1a1, wfaf  &wfaf& normal & float  & \begin{tabular}{c} maxpool (stem) \\ avgpool (intermediate)\end{tabular} \\ \hline
% ReActNet   &  wfaf&w1a1, wfaf  &wfaf& normal  & float  & \begin{tabular}{c} maxpool (stem) \\ avgpool (intermediate)\end{tabular} \\ \hline
BoolNet &  wfaf&w1a1&wfaf& normal  & integer  & \begin{tabular}{c} maxpool (stem) \\ maxpool (intermediate)\end{tabular}\\ \hline
FracBNN    &  w1a1&$\text{w1a1.4}^{\ast}$&w8af& normal& float & \begin{tabular}{c} maxpool (stem) \\ avgpool (intermediate)\end{tabular} \\ \hline
TBN &  wfaf&$\text{w1at}^{{\S}}$ &wfaf& normal & float  & \begin{tabular}{c} maxpool (stem) \\ avgpool (intermediate)\end{tabular}\\ \hline 
PokeBNN    &  w8af&w1a1, w4a4&w8af& \begin{tabular}{c} normal \\ DW, SE \end{tabular}   & float  & \begin{tabular}{c} maxpool (stem) \\ avgpool (intermediate)\end{tabular}\\ \hline
BNext &  w8a8&w1a1, w4a8&w8a8& \begin{tabular}{c} normal \\ SE \end{tabular}   & float  & \begin{tabular}{c} maxpool (stem) \\ avgpool (intermediate)\end{tabular}\\ \hline
ERNs (ours) &  w1a2&w1a2  &w1a2& normal  & integer  & \begin{tabular}{c} avgpool (final)\end{tabular}\\ \Xhline{2\arrayrulewidth}
\end{tabular}
\end{footnotesize}
\end{center}
\end{table*}

%% file: table/ern_result_table.tex
\begin{table}[t]
\caption{ImageNet top1 and top5 accuracy of Efficiera Residual Networks and other ultra-low-bit quantized models whose model size is lower than 10 megabytes. For the unreported model size from original paper, we referred to \citet{zhang2021} and ResNet50 is approximated by assuming input and output layers are float16.}
\label{tab:accuracy}
\centering
\begin{small}
\begin{tabular}{lccc}
\Xhline{3\arrayrulewidth}
Model& \multicolumn{1}{c}{Size (Mbyte)} & \multicolumn{1}{c}{Top1} & \multicolumn{1}{c}{Top5} \\ \hline
Bi-RealNet-18& 4.2 & 56.4 & - \\
Bi-RealNet-34& 5.1 & 62.2 & - \\
QuickNetSmall & 4.0 & 59.4 & - \\
QuickNet & 4.2 & 63.3 & - \\
QuickNetLarge & 5.4 & 66.9 & - \\
HWGQ ResNet18 (w1a2)  & 4.2 & 59.6 & 82.2 \\
HWGQ ResNet34 (w1a2) & 5.1 & 64.3 & 85.7 \\
HWGQ ResNet50 (w1a2) & 7.0 & 64.6 & 85.9 \\
LQ-Nets ResNet18 (w1a2)  & 4.2 & 62.6 & 84.3 \\
LQ-Nets ResNet34 (w1a2) & 5.1 & 66.6 & 86.9 \\
LQ-Nets ResNet50 (w1a2) & 7.0 & 68.7 & 88.4 \\
MeliusNet22 & 3.9& 63.6  & 84.7  \\
MeliusNet29 & 5.1& 65.8  & 86.2  \\
Real-to-Binary Net & 1.9 & 65.4 & 86.2 \\
TBNv2-18 & 4.2 & 59.7 & 82.1 \\
TBNv2-34 & 5.1 & 63.4 & 84.9 \\
TBNv2-50 & 7.0 & 66.6 & 86.7 \\
ReActNet-A & 4.6 & 69.4  & - \\
ReActNet-Adam & 4.6 & 70.5  & - \\
FracBNN & 4.6 & 71.8  & 90.1 \\
PokeBNN-0.5x & 2.0& 65.2  & - \\
PokeBNN-0.75x &  3.8 & 65.2  & - \\
PokeBNN-1.0x & 6.2 & 70.5  & - \\
PokeBNN-1.25x & 9.0 & 73.4  & - \\
BNext-18 & 2.2& 67.9  & - \\
BNext-T & 5.3 & 72.0  & - \\\hline
ERNs18-x0.75 & 0.98 & 63.6 & 85.5 \\
ERNs18   & 1.4 & 65.3 & 86.8 \\
ERNs34   & 2.6 & 70.0 & 89.7 \\
ERNs50   & 3.1 &  72.5 & 91.3 \\
ERNs101  & 5.6 & 73.8 & 92.1 \\ \Xhline{3\arrayrulewidth}
\end{tabular}
\end{small}
\end{table}

%% file: table/gops.tex
\begin{table}[t]
\caption{The number of operations of ERNs and ResNet regardless of the bit-width of weights and activations with 256x256 input resolution.}
\label{tab:gops}
\centering
\begin{scriptsize}
\setlength{\tabcolsep}{4pt}
\begin{tabular}{l|c|c|c|c|c}
\hline
ERNs   & ERNs18-0.75x & ERNs18   & ERNs34   & ERNs50   & ERNs101   \\ \hline
GOPs   & 6.52         & 6.97     & 11.81    & 13.13    & 22.87     \\ \hline
ResNet & -          & ResNet18 & ResNet34 & ResNet50 & ResNet101 \\ \hline
GOPs   &     -         & 4.76     & 9.60      & 10.77    & 20.55     \\ \hline
\end{tabular}
\end{scriptsize}
\end{table}

%% file: table/resolution.tex
\begin{table}[t]
\caption{The effect of the inference scheme to the ImageNet top-1 accuracy. TTA is test-time augmentation with ten-crop augmentation used in \citep{krizhevsky2012, he2016}}.
\label{tab:resolution}
\begin{center}
\begin{tabular}{lrr}
\hline
Inference Scheme   & ERNs-18 & ERNs-50 \\ \hline
224x224 & 60.3  & 68.3 \\ 
256x256 & 62.2  & 69.7 \\
288x288 & 62.8  & 70.3 \\
320x320 & 62.2  & 69.7 \\ \hline
288x288 with TTA & 65.3  & 72.5 \\ \hline
\end{tabular}
\end{center}
\end{table}

%% file: table/fpga.tex
\begin{table}[]
\caption{FPGA resource usage of existing works that achieve ImageNet top1 accuracy over 65.0pt and ours. The number of Block RAM is counted as the number of 18Kb blocks. Notably, Our IP does not consume any DSP blocks. }
\label{tab:fpga}
\setlength{\tabcolsep}{4pt}
\centering
\begin{tiny}
\begin{tabular}{c|c|c|c|c|c}
\hline
Name & LUT  & FF  & Block RAM & Ultra RAM & DSP \\ \hline
\begin{tabular}{c}Synetgy\\\cite{yang2018}\end{tabular} & 51,776 & 42,257 & 318 & N/A & 360 \\
\begin{tabular}{c}T-DLA\\\cite{chen2019}\end{tabular} & 37,921 & 50,508 &  194 &  N/A & 202 \\
\begin{tabular}{c}MobileNetV2\\\cite{wu2019}\end{tabular} & 31,198 & 46,809 &  290 &  N/A & 212 \\
\begin{tabular}{c}JpegComp\\\cite{nakahara2021}\end{tabular} & 274,795 & 273,440 &  2,746 &  16 & 2,370 \\
\begin{tabular}{c}FracBNN\\\cite{zhang2021}\end{tabular} & 50,656 & N/A & 402 & N/A & 224 \\
\begin{tabular}{c} DPU B4096 \\\cite{xilinx2023}\end{tabular} & 52,161 & 98,249 & 510 & 0 & 710 \\
\begin{tabular}{c} Efficiera B64M3C1 \\Ours\end{tabular}  & 77,572 & 83,805 & 280   & 0     & 0   \\ \hline
\end{tabular}
\end{tiny}
\end{table}

% Block RAMB36 132 + RAMB 18 16 =  2*132 + 16 = 280

%% file: table/kv260.tex
\begin{table}[t]
\caption{The inference time and throughput on KV260. The input resolution is fixed to 256x256. The inference time for different input resolution can be estimated by multiplying the relative input area.}
\label{tab:kv260}
\vskip 0.15in
\begin{center}
\begin{scriptsize}
\begin{sc}
\begin{tabular}{lcccr}
\toprule
Model & Inference Time (ms) & FPS (1/s) & Size (Mbyte)\\
\midrule
ERNs18-0.75x & 3.35 & 298.5 & 0.98 \\
ERNs18 & 3.60 & 277.8 & 1.4 \\
ERNs34 & 5.32 & 188.0 & 2.6\\
ERNs50 & 9.64 & 103.7 & 3.1 \\
ERNs101 & 15.78 & 63.4 & 5.6\\
\bottomrule
\end{tabular}
\end{sc}
\end{scriptsize}
\end{center}
\vskip -0.1in
\end{table}